\documentclass{article}

\PassOptionsToPackage{}{natbib}
\usepackage[preprint]{neurips_2025}

\usepackage[utf8]{inputenc} 
\usepackage[T1]{fontenc}    
\usepackage{hyperref}       
\usepackage{url}            
\usepackage{booktabs}       
\usepackage{amsfonts}       
\usepackage{nicefrac}       
\usepackage{microtype}      
\usepackage{xcolor}         
\usepackage{graphicx}

\usepackage{hyperref}
\hypersetup{
    colorlinks=true,
    linkcolor=blue,
    citecolor=teal, 
    filecolor=green,
    urlcolor=blue}

\title{A Fragile Number Sense: Probing the Elemental Limits of Numerical Reasoning in LLMs}

\author{%
  Roussel Rahman\\
  Stanford University\\
  Stanford, CA 94305\\
  \texttt{roussel.rahman@gmail.com} \\
  \And
  Aashwin Ananda Mishra\\
  SLAC National Accelerator Laboratory\\
  Menlo Park, CA 94025 \\
  \texttt{aashwin@stanford.edu} \\
}

\begin{document}

\maketitle

\begin{abstract}
Large Language Models (LLMs) have demonstrated remarkable emergent capabilities, yet the robustness of their numerical reasoning remains an open question. While standard benchmarks evaluate LLM reasoning on complex problem sets using aggregated metrics, they often obscure foundational weaknesses. In this work, we probe LLM mathematical numeracy by evaluating performance on problems of escalating complexity, from constituent operations to combinatorial puzzles. We test several state-of-the-art LLM-based agents on a 100-problem challenge comprising four categories: (1) basic arithmetic, (2) advanced operations, (3) primality checking, and (4) the Game of 24 number puzzle. Our results show that while the agents achieved high accuracy on the first three categories, which require deterministic algorithmic execution, they consistently failed at the number puzzle, underlining its demand for a heuristic search over a large combinatorial space to be a significant bottleneck. These findings reveal that the agents' proficiency is largely confined to recalling and executing known algorithms, rather than performing generative problem-solving. This suggests their apparent numerical reasoning is more akin to sophisticated pattern-matching than flexible, analytical thought, limiting their potential for tasks that require novel or creative numerical insights.
\end{abstract}


Keywords: LLMs; LRMs; Numerical Reasoning; Problem Solving

\section{Introduction}


LLMs have demonstrated exceptional abilities at diverse tasks, such as writing code, composing essays, and solving mathematical problems. A recent evolution is specifically designed for tasks involving reasoning, referred to as Large Reasoning Models (LRMs). However, most of their perceived reasoning abilities are examined using final performance on large benchmark problem sets, while the actual reasoning processes remain less explored. 

LLMs (and LRMs) reason through chains and trees of thoughts, guided by reinforcement learning from sources like human feedback or simulation-based optimization. To understand the reasoning processes, their constituent elements need to be investigated. Importantly, at each step of the chains or the trees, the models reason by probabilistic pattern matching and generate responses as probabilistic token predictions, each with less than 100\% probability of success. Therefore, as the number of reasoning steps increases, the probability of accuracy drops exponentially. Moreover, if the LLMs are weak at one or a few specific steps, the weakness may propagate and cause the trace to diverge towards inaccurate solutions. Evaluations relying on large aggregate benchmarks would obscure this fragility of reasoning that may compromise the model’s reliability and safety in complex environments.

We explore the reasoning abilities of LLMs in the domain of mathematical reasoning and problem solving. A key point to consider is the difference between natural language understanding and understanding mathematical concepts. LLMs are designed to capture patterns in a finite space of words, but the grammar of numbers and the mathematical operations thereon is essentially infinite -- a universe humans explore by exploiting the numerous characteristics of numbers and relationships among them. To illustrate, we consider prime numbers, prime factorizations, transcendental numbers, or the errors introduced when representing them through decimal fractions. Our understanding of these concepts allows us to solve problems of enormous complexity using limited computational resources. We view this human ability to solve problems much more efficiently than permitted by brute-force methods as the core of reasoning abilities -- a view that captures the common theme that runs through several established frameworks of human rationality: bounded rationality \citep{simon1955behavioral}, ecological rationality \citep{gigerenzer2001adaptive}, and resource rationality \citep{lieder2020resource}.

As it is difficult to be efficient in solving mathematical problems purely based on pattern matching and without understanding the concepts, we choose mathematical reasoning as our experimental paradigm. We examine the mathematical reasoning abilities of LLM agents based on their ability to identify the elements of problems (i.e., numbers, operations, and their relationships) and use them to solve complex problems efficiently. We begin our investigation from low levels (e.g., performing operations and calculating expressions) before gradually moving towards higher levels, with a goal to identify where the cracks in reasoning appear and affect complex problem solving performance. As we demonstrate, the LLMs perform reasonably well in the elementary tasks, but they struggle to use these elements in efficiently solving complex problems with uncertainty that require trial-and-error search. Thereafter, we replicate these findings for more advanced LRMs, showing the struggles with efficient search to be a common theme.

\section{Relevant Research on Reasoning}

\subsection{Evaluating LLM Reasoning} 

Reasoning entails making a series of rational decisions using available information and computational resources to achieve specific goals. Mathematical reasoning involves solving math problems by efficiently using mathematical concepts and their relationships. LLM reasoning is typically evaluated via aggregate scores on large benchmarks. For example, the MATH dataset \citep{hendrycks2021measuring} contains over 12500 problems, while the Minif2f dataset \citep{zheng2021minif2f} includes 488 problems from prestigious math competitions (e.g., AIME, AMC, and IMO). Reasoning has also been studied as a component of general intelligence, as in the Abstract Reasoning Corpus for Artificial General Intelligence (ARC AGI) benchmark \citep{chollet2019measure}. While these benchmarks proved challenging for then state-of-the-art agents, LLMs have improved rapidly since \citep{Garisto2022, chollet2024arc}. However, aggregating performance over a large number of problems obscures the sources of improvement, leading researchers to seek more granular explanations of LLM reasoning.

\subsection{Explaining LLM Reasoning}

\citet{wei2022chain} showed that \textit{Chain of Thought} (CoT) prompting -- simply asking models to divide complex problems into smaller subproblems and reason step by step -- considerably improves reasoning performance without additional training. The self-reported steps serve as explanations of reasoning. Extending CoT, \citet{yao2023tree} introduced the \textit{Tree of Thought} (ToT) approach that represents problems as trees of subproblems before systematically searching through numerous reasoning paths. Recent efforts integrate tree-search methods with reinforcement learning from human feedback \citep[e.g., ][]{bai2022training} or self-reflections \citep[e.g.,][]{ouyang2022training}. CoT, ToT, and reinforcement learning have become integral elements of modern LLM agents' reasoning abilities.

Despite their usefulness, the LLM self-reports have been questioned as to whether they reflect the true processes used in reaching solutions \citep{ahn2024large, turpin2023language}. To specify the details of the reasoning processes, Explainable AI (XAI) offers several classes of tools of varying granularity \citep{bereska2024mechanistic}. A particularly promising approach has been mechanistic interpretability (MI), which seeks to explain behavior by reverse-engineering MLP-based AI models from neuron-level information. For example, \citet{bricken2023towards} identified interpretable features corresponding to distinct reasoning operations from neural activations. \citet{nanda2023progress} reverse-engineered the algorithm learned by a single-layer transformer for modular addition (i.e., problems of the form $mod(a+b, n)$). While MI has been useful in deciphering LLM reasoning, scaling for more complex problems remains a challenge \citep{bereska2024mechanistic}. 

\citet{shojaee2025illusion} examined how problem complexity affects reasoning by designing controllable puzzle environments -- such as the Tower of Hanoi (ToH) and river crossing tasks -- that allow complexity to vary while keeping the environment constant. Their findings reveal that although LRMs appear to ``think'' better, their reasoning remains fragile. On low-complexity problems, standard LLMs outperformed LRMs; on medium-complexity tasks, LRMs had the edge. Yet both failed on high-complexity problems, underscoring the limits of LLM reasoning. The collapse in ToH performance is particularly surprising, given the existence of deterministic optimal strategies that apply across all complexity levels. Humans tend to discover these strategies rapidly through learning (see Section \ref{sec:human_reasoning}). Moreover, \citeauthor{shojaee2025illusion} observed that the LRMs sometimes overlooked the right solutions and continued towards worse solutions, suggesting a lack of basic problem understanding. A primary goal of our work is to assess the LLMs' understanding of mathematical elements as a step toward comprehending their overall problem-solving processes.

\subsection{Explaining Human Reasoning} \label{sec:human_reasoning}

The ToH task, with its vast search space, challenges new learners but also contains discoverable subgoals that can simplify the problem. \citet{anzai1979theory} investigated how an individual unfamiliar with the ToH task discovered new strategies as they learned to solve a five-disk version over four trials. Their strategy evolved in each attempt -- from selective search, to goal-peg planning using means-end analysis, then to a recursive subgoal strategy, and finally to a pyramid strategy that treated several disks as sets. Notably, each of the last three strategies yields optimal performance, but the latter strategies become more efficient to represent and execute. Previously, \citet{simon1976strategy_shift} examined the strategy shifts in a variant of the river crossing problem. They showed that in such complex problems, an ``all-or-none" approach in search of optimality is impossible; rather, we need to use heuristics to find acceptable solutions that satisfy and suffice for our goals within limited resources. This principle forms the basis of the bounded rationality in humans that enables us to solve complex problems using limited computational resources \citep{simon1955behavioral, newell1958elements, simon1971human, gigerenzer2001adaptive, gigerenzer2020bounded}.

Specific to mathematical reasoning, we possess an abstract number sense -- an intuitive grasp of numbers, their relationships, and manipulations -- that enables us to quickly estimate and reason without formal calculations \citep{dehaene2001precis}. Its origins have been explored from multiple perspectives. \citet{gallistel1992preverbal} described two systems: a preverbal one for approximating quantities early in development and a verbal one for precise calculations shaped by cultural learning. \citet{dehaene1992varieties} proposed a triple code model: visual numerals (e.g., ``37''), verbal labels (``thirty seven''), and analog magnitudes distributed along a number line. While number sense is abstract, neuroimaging studies reveal a specific brain region (the inferior parietal cortex) that contributes heavily to it \citep{dehaene1996organization, kiefer1997time, pinel1999event}. A part of number sense is biologically determined and has evolutionary roots, and a part is learned and developed. 

Generally, studies of human reasoning and developmental changes sought boundedly rational explanations through processes or strategies that are implementable within available resources, as opposed to a black box relating input information to output performance. For example, while learning addition, children have been observed to adaptively use at least five different strategies \citep{siegler1984strategychoices, siegler1987perils} and their strategy transitions with learning have been emulated in cognitive models \citep{shrager1998scads}. For the ToH, \citet{anzai1979theory} predicted the sequence in which a learner would discover improved strategies. For an accurate understanding of LLM reasoning, especially in mathematical problems, such process-level explanations are essential.

\section{Objectives, Framework, \& Application} 

Do LLMs acquire a true \textit{number sense}? Human numerical intuition allows us to simplify complex problems and navigate vast solution spaces efficiently. Without this sense, LLMs may excel at executing deterministic calculations but are likely to falter when tasks demand heuristic search, such as in trial-and-error problem-solving. This makes mathematical challenges an ideal paradigm for investigating how LLMs integrate elementary skills into complex strategies. While some mechanistic studies show that LLMs encode basic mathematical relationships, how these are integrated for higher-order reasoning remains less explored. This is partly because existing benchmarks and interpretability methods often face a trade-off between task complexity and analytical specificity \citep{wang2022interpretability, bricken2023towards, bereska2024mechanistic}.

\textbf{Approach: A "Divide-and-Reconstruct" Framework.} We propose a multi-level framework to enable scalable and interpretable analysis of LLM numerical reasoning. The core idea is to connect high-level task performance with its low-level building blocks. Our long-term vision is to use this framework to identify persistent reasoning traits (e.g., recurring error patterns) and trace their causal roots across layers of analysis: from observable behavior (task success/failure, accuracy), to self-explanation (the model's rationale), and finally to internal computation (model internals). Inspired by cognitive science and mechanistic interpretability, we adopt a ``divide-and-reconstruct'' strategy: first, we decompose a complex task into its elementary skills, and then we evaluate LLMs on both the isolated skills and their integration.

\textbf{Application: Probing Numerical Reasoning.} As an application of this framework, we investigate the limits of LLM numerical reasoning. We focus on the known challenge of problems requiring heuristic search \citep{yao2023tree}. Following our "divide-and-reconstruct" strategy, we first test LLMs on elementary mathematical operations. We then examine their ability to integrate these skills in two progressively complex tasks: (a) primality checking, which can be solved by a deterministic search algorithm (i.e., trial division) with a variable number of steps, and (b) the Game of 24, which requires a non-deterministic, combinatorial search. This progression allows us to trace how foundational numerical skills (or their fragility) contribute to success or failure on more complex problems.

\section{Experimental Methods}


\subsection{Problem Types}

Our test contained 100 problems divided equally into four sets by problem type. 

\textbf{Set 1: Basic Mathematical Operations} This set allows us to set a reference for LLM mathematical skills in problems that involve only four basic mathematical operations -- addition, subtraction, multiplication, and division -- and require minimal reasoning. We begin with simple problems (e.g., 2 + 2 + 4 - 2 = ? and 1/13 = ?) before increasing difficulty by involving longer chains of operations. Some problems involve transcendental numbers (e.g., $\pi$ and $e$) and reciprocals of primes to exploit a numerical precision problem observed in LLMs.

\textbf{Set 2: Advanced Mathematical Operations}
We introduce more advanced operations and concepts, such as exponentiation, logarithms, and complex numbers. Most problems require combinations of operations. Some problems contain slightly unusual values, such as fractions as exponents (e.g., $400^{0.23}$) and logarithms with base 5 or 13 (e.g., $\log_{13}{169}$). This set provides a sterner test of LLM abilities in executing chains of operations than Set 1, but still offers a minimal scope of reasoning.

\textbf{Set 3: Check if a Number is Prime}
We ask the agents to check the primality of numbers. We varied the length of the numbers (discontinuously from 1- to 54-digit numbers). Some numbers are relatively famous and likely to be discussed in LLM training data, such as Mersenne Primes of the form $2^p-1$ (e.g., $2^{13}-1$ and $2^{31} - 1$). We also include some non-primes of the same form ($2^{27}-1$ and $2^{29}-1$), which we refer to as the \textit{Mersenne Prime Lookalikes}. 


This set allows us to examine LLM performance in combining basic skills to solve a more complex problem. The algorithms for checking the primality of numbers by searching for factors are well-known; however, the point at which to stop searching is non-deterministic. There are explicit clues in (prime) number theory to exploit -- such as any number of the form $2^p-1$ cannot be prime if $p$ is not prime (i.e., $2^{27} -1$ is not prime) -- or more implicit ones in our number sense (e.g., we can tell $333$ or $77777$ are not prime without explicit reasoning).

\textbf{Set 4: The 24 Game} $24$ is a popular number game that has been previously used in teaching children and testing LLMs \citep{van2017cognitive, ding2023everything}. The game requires only elementary mathematical knowledge to play. A player is given a set of four numbers (e.g., $[2, 4, 6, 6]$ and their goal is to produce $24$ as the output, using each number once, and the four basic operations -- addition, subtraction, multiplication, and division -- as many times needed (e.g., $(6+2-4) \times6 = 24$). 

Despite its simplicity, the search space is vast with numerous combinations of numbers and operations to consider. This space can be reduced using number relationships as clues. For example, a helpful strategy is to look for factors that make up $24$ or for multiples of $24$. In the above example, we may set aside one of the $6s$ and try to manipulate the rest of the numbers to get the required $24/6=4$.
There are numerous such strategies, but none of them guarantee success; rather, solvers need to find the solutions through an informed trial-and-error search using their ``number sense."

\subsection{Participants}

We began with eight LLMs and ran an initial test of competence on basic mathematical operations for inclusion in the main test. Three models -- ChatGPT 4o, Cohere AI, and Mistral AI -- were excluded as they scored below our threshold of 80\% accuracy, and the remaining five (Table \ref{table:models_study1}) proceeded to the main study. To validate our findings, we later ran a follow-up test with three models (Table \ref{table:models_study2}).


Notably, the agents differ substantially in architecture, and the differences further propagate due to fine-tuning for specific abilities. For example, both o1 models are equipped with enhanced reasoning, but the o1 model ``thinks" longer using more chains than the o1-mini. The remaining three models were trained for general-purpose usage. Complete details are unavailable due to the proprietary nature of the models. Some high-level similarities are: (1) all use transformer models pre-trained on large corpora of texts, and (2) follow probabilistic principles for token generation using contextual information captured by positional encodings, self-attention mechanisms, and multi-head attention.

\begin{table}[t]
\caption{Five Models included in the \textbf{main study} and their versions. These models were tested over a 7-day period between December 30, 2024 -- January 5, 2025 PST.}
\label{table:models_study1}
\vskip 0.15in
\begin{center}
\begin{small}
\begin{sc}
\begin{tabular}{lccr}
\toprule
Model & Version & Reasoning\\ & & Enabled? \\
\midrule
OpenAI ChatGPT    & o1-mini    			& Yes 				\\ 
OpenAI ChatGPT    & o1         			& Yes 				\\ 
Google Gemini     & 1.5        			& No 				\\ 
Anthropic Claude  & Sonnet 3.7 			& No 				\\ 
Microsoft Copilot & -          			& No 				\\
\bottomrule
\end{tabular}
\end{sc}
\end{small}
\end{center}
\vskip -0.1in
\end{table}

\begin{table}[b]
\caption{Three Models included in the \textbf{follow-up study} and their versions. These models were tested over a 5-day period between June 16, 2025 -- June 20, 2025 PST.}
\label{table:models_study2}
\vskip 0.15in
\begin{center}
\begin{small}
\begin{sc}
\begin{tabular}{lccr}
\toprule
Model & Version & Reasoning\\ & & Enabled? \\
\midrule
OpenAI ChatGPT    & o3    			& Yes 				\\ 
Google Gemini    & 2.5 Pro        	& Yes 				\\  
DeepSeek          & R1 (DeepThink)  & Yes 				\\
\bottomrule
\end{tabular}
\end{sc}
\end{small}
\end{center}
\vskip -0.1in
\end{table}

\subsection{Experimental Procedure and Scoring}

In the main study, we presented the agents with six prompts in sequence (Appendix \ref{sec:prompts_used}). First, we explained that they would be provided with some problem sets to solve and asked to show their steps before summarizing answers in ordered lists. In prompts 2-4, we presented the first three problem sets with brief task explanations. In prompt 5, we asked the agents about the 24 game and its standard rules. After confirming they had the intended version in mind, we gave them 25 games to play. 

To examine the effects of complexity on reasoning in the 24 game, we conducted a follow-up study, creating a second, harder set of 24 games and contrasting LLM performance from the easier set used earlier. Here, we used two prompts in sequence (Appendix \ref{sec:prompts_used_study2}). In the first prompt, we asked the LLMs to state the game rules and provided the same set of 25 games as in the main study (Set 4). In the second prompt, we provided 25 additional games of higher difficulty. As a notable difference from the first study, we explicitly instructed in the prompts not to use any coding tools to prevent solutions through exhaustive search, so that their ability to solve the problems efficiently through reasoning could be examined. The prompts proved sufficient to prevent the Gemini and Deepseek models from coding, but we needed to disable the coding abilities for the ChatGPT model.

We scored each correct answer by 1 point and a wrong answer by 0. For decimal fractions as answers, we asked for three digits after the decimal point, but we awarded the point if the first two digits were correct. Each agent performed the test three times, and the scores were averaged over three trials.

\section{Results}

\subsection{Overall Test Performance of LLMs}

\begin{figure}[!t]
    \centering
    \includegraphics[width=0.45\columnwidth]
    {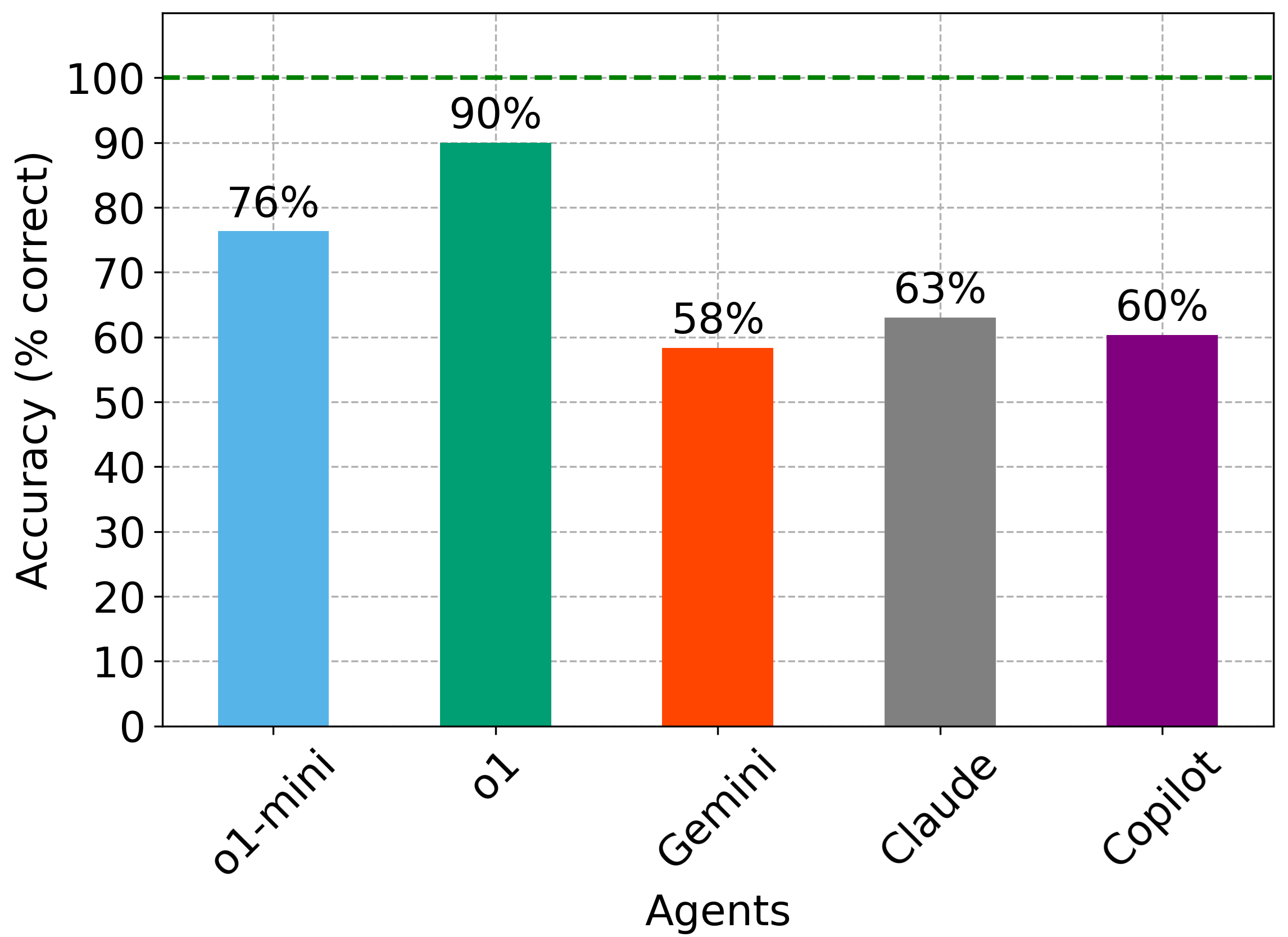}
    \caption{Overall performance of the models represented by the average score across problem sets.}
    \label{fig:overall_scores}
\end{figure}

The overall performance of the agents is shown in Figure \ref{fig:overall_scores}. ChatGPT o1 model performed the best (90\% average accuracy), followed by o1-mini (76\%). The remaining three models performed at a level similar to each other (58-63\%) but considerably poorer than the two ChatGPT models. The performance discrepancy is not surprising since the ChatGPT models are equipped with explicit reasoning abilities. 
However, this high-level view obscures their competence in specific mathematical tasks. Consequently, we cannot determine which 10\% of the problems the o1 agent made mistakes in, or whether Claude, Gemini, and Copilot made mistakes in the same problems, despite their similar scores. In the following two sections, we focus on specifying the scope of their mistakes.

\subsection{LLM Performance in Different Problem Types}

\begin{figure*}[!b]
    \includegraphics[width= 0.98\textwidth]
    {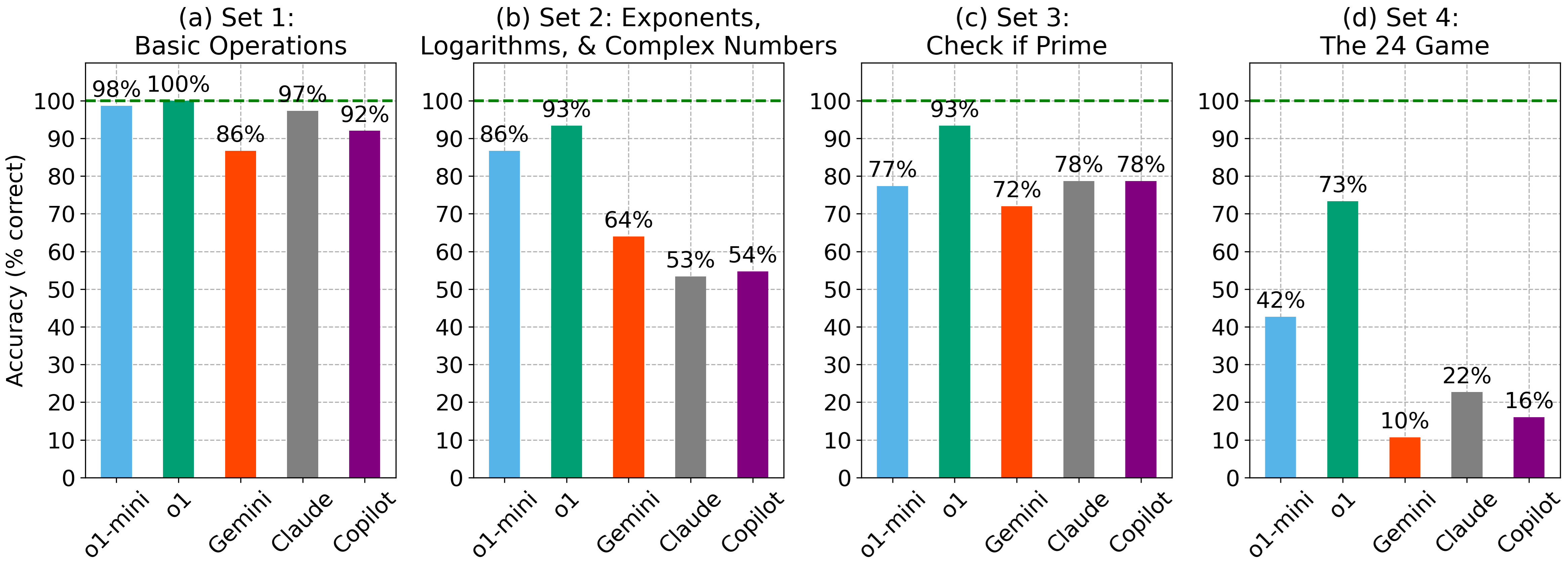}
    \caption{Performance of each agent in each of the four problem categories. The agents performed reasonably well in the first three sets but struggled in the fourth set, which contained the games of 24.}
    \label{fig:performance_by_category}
\end{figure*}

Figure \ref{fig:performance_by_category} shows LLM performance divided across problem types. All agents performed well on the basic operations (Set 1, Fig. \ref{fig:performance_by_category}a), with the lowest accuracy being 86\%, showing all agents included in the test possess the elementary mathematical skills needed to perform the other tasks in the test. Therefore, the scores in the other tasks reflect how well they were able to generalize the elementary skills needed to solve more complex problems.

On Set 2 (Fig. \ref{fig:performance_by_category}b) containing advanced operations, the two ChatGPT models performed reasonably well, both scoring above 80\%, with only a slight drop from the first set. On the other hand, the scores of Gemini, Claude, and Copilot agents drop considerably. These three agents scored between 53-64\%, in contrast to their above 85\% average scores on the first set.

On Set 3 (Fig. \ref{fig:performance_by_category}c), we examined the agents' ability to use the knowledge of factors (tested on Set 1) to check if the given numbers are prime. As we can see, the o1 model retained its superior performance level on Set 3. In contrast, the o1-mini agent dropped down to the level of the other three agents, who slightly raised their performance from the advanced operations set.

On Set 4 (Fig. \ref{fig:performance_by_category}d) containing the games of 24, the performance of all LLM agents dropped sharply compared to the first three sets. To provide a comparison, on average, all agents scored between 74\% and 95\% on the first three sets, whereas the agents scored between 11\% and 73\% on Set 4. The o1 model is the top performer at 73\%, whereas all other agents solved less than 50\% of the problems.

These results show that while the agents perform reasonably well in fundamental operations and following deterministic steps to solutions (as reflected in the first three tasks), they struggle to transfer these skills to solve the games of 24. Notably, the 24 game does not require any new operations; rather, the basic operations tested on Set 1 suffice for this game. However, unlike the first three tasks, the 24 game also requires a trial-and-error search for solutions, which has been noted as a challenge for LLMs. In the next section, we examine the mistakes made in each problem type to specify the points of their struggles.

\subsection{A Closer Look at the Types of Mistakes}

\begin{figure*}[!b]
    \includegraphics[width=0.98\textwidth]
    {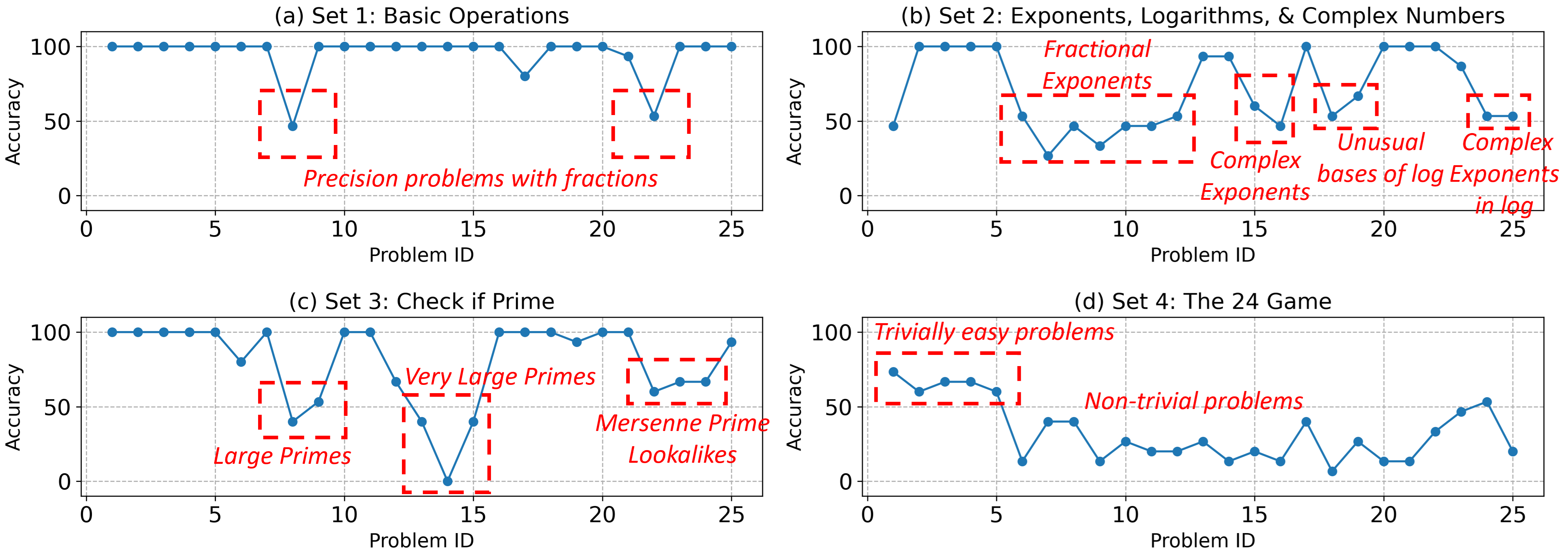}
    \caption{Performance on individual problems. As can be seen, some problems were harder than the rest for the agents. Annotations include problem features as the possible reasons for observed errors.}
    \label{fig:mistakes}
\end{figure*}

Average LLM performance in each problem is presented in Figure \ref{fig:mistakes}. 
On Set 1, all LLMs can be observed to solve most problems perfectly. The mistakes were mainly limited to loss of precision in decimal fractions. These mistakes could be avoided by symbolic operations to simplify the problems, a strategy often adopted by the agents. On Set 2, the LLMs struggled with multiple types of operations. All LLMs -- barring the ChatGPT agents -- struggled with the fractional exponents (such as $400^{0.23}$). The agents also made more mistakes in problems involving unusual bases of logarithms and complex numbers. On Set 3, the LLMs performed generally better than they did on Set 2, but again showed some pitfalls. The most common mistake predictably coincided with large prime numbers. In contrast, they were unaffected by the size of the numbers when dealing with Mersenne primes, but they often mistook the Mersenne Prime Lookalikes (i.e., the non-primes of the form $2^p-1$) for prime numbers.

On Set 4 containing the games of 24, LLM performance can be observed to fall off a cliff after the initial few problems that were trivially easy and could be solved by simply multiplying the given numbers (such as [1,1,1,24] or [1,1,3,8]). Surprisingly, some agents were not able to solve even these problems. The rest of the problems were more complex than that but still relatively easy for humans (presented in Appendix \ref{sec:prompts_used}), but all LLM agents showed imperfect performance in these problems. A common mistake was assuming a solution did not exist for many of the problems, whereas all problems contained at least one solution. Moreover, the agents frequently broke the rules of the game to achieve 24, mainly in two ways: (1) not using all the numbers and (2) using some numbers more than once. Finally, another common set of mistakes was miscalculating expressions and then providing solutions that do not yield 24 without realizing the miscalculations. These mistakes are particularly surprising as they indicate a breakdown in basic mathematical skills, which we had observed the agents to possess when tested on simpler problems. Notably, the ChatGPT models did not make the last kind of mistakes and performed generally better in playing the 24 game.

\subsection{Examining the Effect of Complexity on Reasoning}

\begin{figure}[!t]
    \centering
    \includegraphics[width=0.50\columnwidth]
    {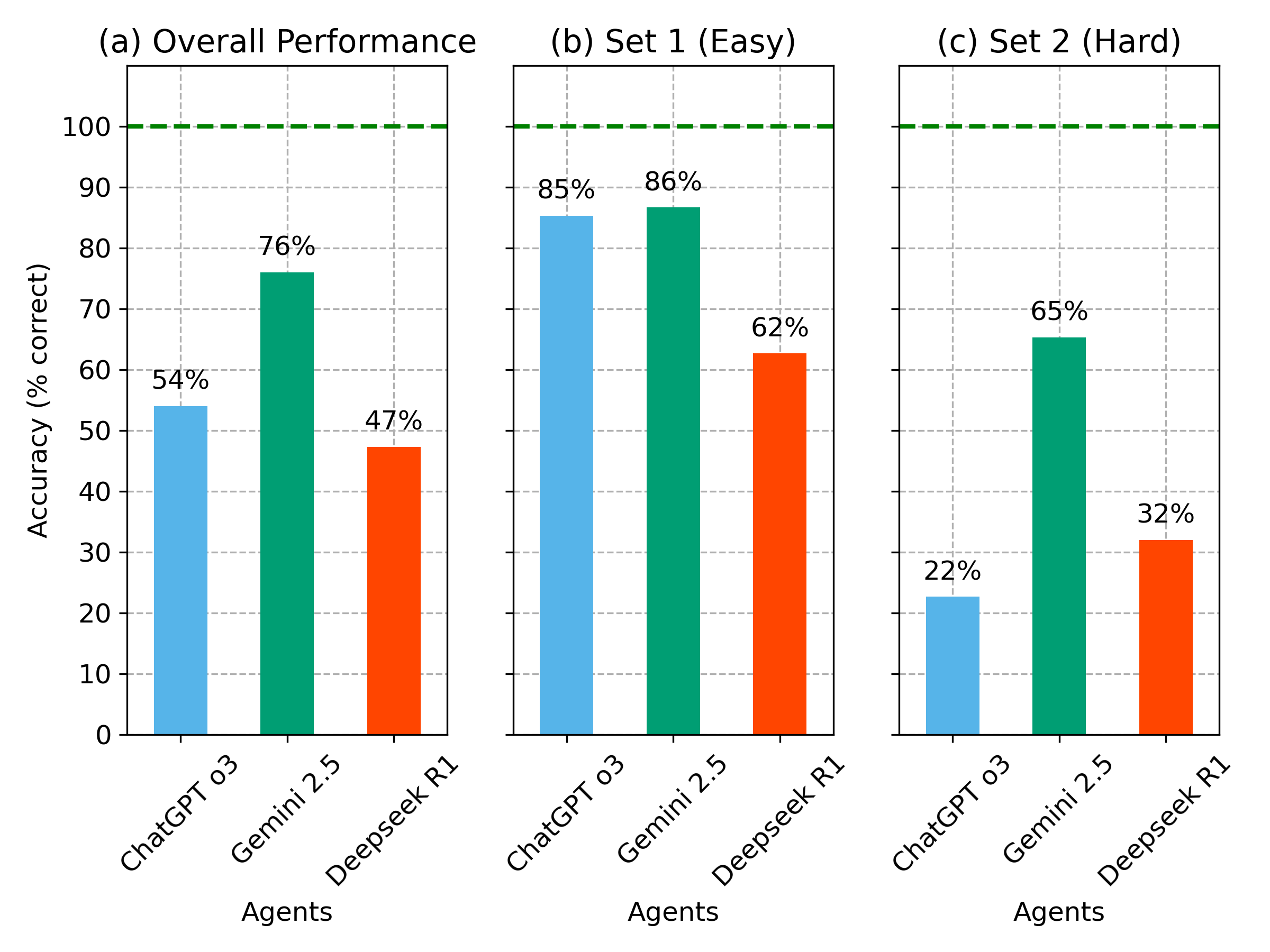}
    \caption{Performance of the three LRM agents in the game of 24 at two levels of difficulty.}
    \label{fig:performance_june_study}
\end{figure}

Figure \ref{fig:performance_june_study} shows the performance of the three LRM agents in the 24 games. Figure (a) shows the average performance across two levels of difficulty. As we can see, Gemini 2.5 Pro agent performed the best (76\%), comfortably outperforming both the ChatGPT o3 (54\%) and the Deepseek R1 (47\%) agents. Figures (b) and (c) show the performance broken down by problem difficulty level. On Set 1, which is relatively easier, the ChatGPT o3 and the Gemini 2.5 Pro agents perform equally well -- 85\% and 86\% accuracy, respectively -- while the Deepseek R1 scored 62\%. On Set 2, containing harder games of 24, we see that all agents perform considerably worse compared to the first set. The Gemini agent's score drops down to 65\%, whereas the ChatGPT and the Deepseek agents suffer even larger dips in performance, with their scores going down to 22\% and 32\%, respectively. These results suggest that the trial-and-error search required in the 24 game is a persistent challenge for modern LLMs, despite enhanced reasoning abilities.

Examining the type of mistakes made, we find that erroneously assuming a solution does not exist remains the most common mistake. Moreover, one such conclusion increased the tendency to conclude the same for later problems. On the other hand, we find that these LRMs were less prone to breaking the rules of the game (e.g., not using all numbers exactly once), with the Gemini model making the fewest such mistakes and the Deepseek model making the most. The frequency of miscalculating expressions also decreased compared to the models tested in Study 1. However, on the harder set, all agents sometimes diverged towards endless chains of reasoning, during which the number and the frequency of miscalculating expressions appeared to increase, indicating a possible effect of compounding errors.

\section{General Discussion}

As LLM-based agents become our assistants on complex, real-world tasks, understanding their complex reasoning abilities has become essential. In this work, we tested LLM reasoning in the domain of mathematical problem solving as they navigated vast spaces of numbers to find solutions. In our main study, we tested five leading agents on a set of 100 problems across four categories: basic operations, advanced operations, primality checking, and a number game. The agents performed reasonably well in the first three categories, followed by a sharp drop in the number game. This result suggests that while LLMs can effectively execute deterministic problem-solving strategies, they struggle with tasks that require "number sense" -- an intuitive understanding of number characteristics and relationships that humans naturally possess and further develop through learning. This limit becomes evident in the number game, which requires an informed trial-and-error search using knowledge of number factors to guide problem-solving attempts.

More recent tests on new reasoning models indicate that the 24 game and the trial-and-error search inherent in it are persistent challenges for LLMs. While reasoning models perform better on hard problems than non-reasoning models, their improvement comes at the cost of increased computation, as they often embark on long chains of reasoning or code execution. The trade-off between reasoning capability and computational efficiency suggests the need for evaluative frameworks to consider both problem complexity and solution costs for practical applications. Moreover, despite additional computational capabilities, their performance dropped on the harder set of 24 games. Therefore, their superiority over non-reasoning LLMs seems to stem primarily from increases in computational resources rather than improved number senses.

Probing deeper, the types of errors made reflect a fragile number sense in LLMs. Erroneously assuming the non-existence of solutions is the most common mistake, followed by breaking the game rules and miscalculating expressions. An interesting phenomenon we observed is the breakdown of elementary mathematical skills within extended chains of reasoning. While all models included were capable of performing basic operations well, their skills did not always transfer to reasoning chains. They frequently miscalculated simple expressions in producing 24 and, except for the LRMs, were rarely aware of their mistakes.

Taken together, our findings highlight a fragile number sense in LLM agents, despite their skills in following deterministic processes. This limited number sense is likely to make trial-and-error search in vast number spaces a persistent challenge for LLMs. More generally, these results highlight the need for computationally limited agents to be boundedly rational in complex environments, that is, to selectively search for good-enough solutions using heuristics when optimality is beyond reach. As number sense allows humans to solve complex mathematical problems efficiently, we believe it to be essential for improving the efficiency LLM reasoning. The findings from our relatively simple study show the merits of the divide-and-reconstruct approach to decipher how intelligent agents combine the low-level elements of reasoning to solve problems of immense complexity.



\subsection*{Limitations and Future Directions}

To establish trial-and-error search in complex environments and, more fundamentally, number sense as limits of LLM mathematical reasoning, they need to be replicated in larger sets of tasks at varying levels of complexity. Some promising candidates include number puzzles (24, four fours, or Sudoku), Cryptoarithmetic problems, and synthetic number problems. Alternatively, problems may be chosen based on their computational complexity (e.g., NP-Hard class of problems). Moreover, we examined only a few selected LLMs among the vast array available today. Even across variants of the same models, the agents can differ drastically in abilities and performance. 
Therefore, future studies need to examine a larger, diverse pool of models. Finally, as complex problems allow numerous solving strategies, the paths from input information to output performance need to be traced for causal attribution. In this pursuit, XAI methods such as MI reflect the cognitive models used to explain human behavior as promising ways to explain AI reasoning from its elementary processes. Our study highlights several common mistakes, such as producing non-24 outcomes, breaking 24 game rules (e.g., reusing numbers or omitting some), and the breakdowns of reasoning chains, that MI could unpack by identifying the building blocks (e.g., features, circuits, and motifs) involved in reasoning and examining how these internal representations drift during extended reasoning.

\section{Conclusions}
Our examination of the mathematical reasoning abilities of LLM-based agents revealed clear strengths in deterministic operations such as basic arithmetic, advanced calculations, and prime number checks. However, performance dropped sharply on tasks requiring trial-and-error reasoning, such as the 24 game. These outcomes underscore a significant gap between human and LLM performance in managing uncertainty and solving complex problems. More generally, our study highlights the merits of conducting targeted experiments to systematically evaluate and explain LLM abilities in navigating complex spaces. Further explorations of this nature may pave the way for improved AI safety by revealing ways to improve large LLM reasoning, as well as by drawing clearer distinctions between AI and human reasoning in complex real-world problems.

\begin{ack}
We are sincerely grateful to Michael J. Schoelles and Jeff Shrager for sharing their vast knowledge of mathematical reasoning and providing invaluable guidance in this research.
\end{ack}

\section*{Code and Data Availability}
The problem sets in full are included in Appendices \ref{sec:prompts_used} and \ref{sec:prompts_used_study2}. We highly encourage readers to test their LLM assistants on these problems, especially the games of 24. The complete data generated in this work can be found here: \url{https://osf.io/fncyu/}.

\bibliographystyle{apalike}
\bibliography{bibliography}








\newpage
\appendix
\section{Prompts Used in Study 1}\label{sec:prompts_used}


\noindent\underline{Prompt 1}: Hi. I will give you some sets of mathematical problems. Solve them as best as you can. For any fractions, provide 3 digits after the decimal point. You are welcome to provide the reasonings, but please remember to summarize all answers in a list following the order of the problems presented. Do you understand?\\

(After an affirmative response from the agent)\\

\noindent\underline{Prompt 2}: Here is the first set of problems. Simply calculate these values.

\begin{enumerate}
	\item $1+2-3*4/5$
	\item $2+2+4-2$
	\item $(3+4*3)/5$
	\item $1/13$
	\item $13*7/(169*25)$
	\item $(13+4)(13-4)$
	\item $1/1.2+1/6$
	\item $(1/13-1/11)*999999$
	\item $87/29+2$
	\item $49\pi$
	\item $\pi/6$
	\item $e+\pi$
	\item $79/13$
	\item $1+2+3+4+5+6+7$
	\item $1-2+3-4+5-6+7-8$
	\item $1-1+1-1+1$
	\item $55 -1-2-3-4-5-6-7-8-9$
	\item $1+1/2+1/3+1/4+1/5+1/6+1/7$
	\item $1/10+1/100+1/1000+1/10000$
	\item $11*101/9999$
	\item $1/(11*101*73*137)$
	\item $(11*13*17*19)/(22*39*68*95)$
	\item $1/(3/10)-20/6$
	\item $1/(2*12-3*(73-69))$
	\item $1/(2*12-2*3*(73-69))$
\end{enumerate}
\underline{Prompt 3}: Okay. Now, calculate these values.

\begin{enumerate}
	\item $0^0$
	\item $1^0$
	\item $2^3$
	\item $2^{-5}$
	\item $400^{-.5}$
	\item $400^{0.23}$
	\item $237^{3.7}$
	\item $500^{2/7}$
	\item $300^{3.1416}$
	\item $700^{-1/7}$
	\item $71^{0.26}$
	\item $4900\pi^2$
	\item $(\pi +2)(\pi-2)-\pi^2$
	\item $\pi/10 + log(e^{\pi/5})$
	\item $71^{0.26e^{\pi i}}$
	\item $71^{0.26e^{2\pi i}}$
	\item $log_2 (2^{0.23})$
	\item $log_{23} (0.309)$
	\item $log_\pi((1+i)^2-2i)$
	\item $log_8 (512)$
	\item $log_5(1/625)$
	\item $log_5(1/625)$
	\item $log_2 (2^{20000000000000000000000000000000000000})$
	\item $log_2 (2^{20000000000000000000000000000000000000i})$
	\item $log_{13}(999999/(7*27*37*11))$
\end{enumerate}

\noindent\underline{Prompt 4}: Now, check if these numbers are prime or not.

\begin{enumerate}
	\item $5$
	\item $233$
	\item $349$
	\item $361$
	\item $367$
	\item $499$
	\item $71993$
	\item $1282529$
	\item $3326489$
	\item $514229$
	\item $91193$
	\item $99194853094755497$
	\item $1066340417491710595814572169$
	\item $263130836933693530167218012159999999$
	\item $359334085968622831041960188598043661065388726959079837$
	\item $2^7-1$
	\item $2^{13}-1$
	\item $2^{17}-1$
	\item $2^{19}-1$
	\item $2^{31}-1$
	\item $2^{11}-1$
	\item $2^{23}-1$
	\item $2^{29}-1$
	\item $2^{37}-1$
	\item $2^{27}-1$
\end{enumerate}

\underline{Prompt 5}: Great! The next set is a bit different. Are you familiar with the 24 game and its rules? If you are, I will give you 25 games to play (i.e., 25 sets of 4 digits to make 24 with).\\

(After an affirmative response from the agent and verifying that they know the rules)\\

\underline{Prompt 6}: Perfect. You are welcome to provide the reasonings, but please remember to summarize all answers in a list following the order of the problems presented. Here is a set of 25 problems.

\begin{enumerate}
	\item $1,1,1,24$
	\item $1,1,2,12$
	\item $1,2,3,4$
	\item $1,1,4,6$
	\item $1,1,3,8$
	\item $0,15,39, 81$
	\item $3,3,4,5$
	\item $2,2,2,8$
	\item $6,6,8,12$
	\item $1,3,3,3$
	\item $4,5,9,13$
	\item $1,1,1,12$
	\item $5,9,12,13$
	\item $6,6,8,8$
	\item $4,6,7,9$
	\item $7,8,9,12$
	\item $2,4,6,6$
	\item $6,7,8,12$
	\item $1,2,3, 13$
	\item $3,9,12,13$
	\item $1,1,6,12$
	\item $1,5,6,12$
	\item $1,11,12,13$
	\item $2,2,8,8$
	\item $2,8,8,9$
\end{enumerate}

\newpage
\section{Prompts Used in Study 2}\label{sec:prompts_used_study2}

\underline{Prompt 1}: Are you familiar with the 24 game and its rules? If you are, I have 25 games for you to play (i.e., 25 sets of 4 numbers to make 24 with). First, state the rules briefly, and then start playing! One strict condition, though: You cannot write or use any code at all, even during your ``thinking" processes. That would mean direct disqualification from the game.

\begin{enumerate}
	\item $1,1,1,24$
	\item $1,1,2,12$
	\item $1,2,3,4$
	\item $1,1,4,6$
	\item $1,1,3,8$
	\item $0,15,39, 81$
	\item $3,3,4,5$
	\item $2,2,2,8$
	\item $6,6,8,12$
	\item $1,3,3,3$
	\item $4,5,9,13$
	\item $1,1,1,12$
	\item $5,9,12,13$
	\item $6,6,8,8$
	\item $4,6,7,9$
	\item $7,8,9,12$
	\item $2,4,6,6$
	\item $6,7,8,12$
	\item $1,2,3, 13$
	\item $3,9,12,13$
	\item $1,1,6,12$
	\item $1,5,6,12$
	\item $1,11,12,13$
	\item $2,2,8,8$
	\item $2,8,8,9$
\end{enumerate}

\underline{Prompt 2}: Perfect! Now, solve these 25 problems. Remember the condition: You cannot write or use any code at all, even during your ``thinking" processes. That would mean direct disqualification from the game.

\begin{enumerate}
	\item $7,10,12,13$
	\item $4,4,8,9$
	\item $2,8,10,12$
	\item $2,3,9,12$
	\item $5,9,12,13$
	\item $3, 5, 9, 10$
	\item $2,5,7,8$
	\item $1,4,9,13$
	\item $5,9,10,11$
	\item $2,3,8,13$
	\item $5, 10, 10, 13$
	\item $2,3,8,13$
	\item $3,5,8,13$
	\item $3,5,9,10$
	\item $4,5,6,12$
	\item $2,2,3,5$
	\item $2,4,7,8$
	\item $3,4,5,11$
	\item $4,4,7,7$
	\item $2,7,7,10$
	\item $2,5,10,12$
	\item $3,9,9,11$
	\item $2,9,13,13$
	\item $2,3,9,12$
	\item $5,7,9,10$
\end{enumerate}

\end{document}